\documentclass[11pt]{article}

\usepackage[final]{acl}

\usepackage{times}
\usepackage{latexsym}
\usepackage{amsmath}
\usepackage{amssymb}
\usepackage{booktabs}
\usepackage{multirow}
\usepackage{xcolor}

\usepackage[T1]{fontenc}
\usepackage[utf8]{inputenc}

\usepackage{microtype}

\usepackage{inconsolata}

\usepackage{graphicx}

\title{Robust LLM Data Selection via Content-Invariant Gradient Attribution}

\author{
 \textbf{Jia Gu\textsuperscript{1,2}},
 \textbf{Liang Pang\textsuperscript{1,2}\thanks{*Corresponding author}
},
 \textbf{Huawei Shen\textsuperscript{1,2}}
\\
 \textsuperscript{1}Institute of Computing Technology, Chinese Academy of Sciences, Beijing, China \\
 \textsuperscript{2}University of Chinese Academy of Sciences, Beijing, China
\\
 \small{
   \textbf{Correspondence:} {{gujia24s,pangliang,shenhuawei\}@ict.ac.cn}
}
}
}

\begin{document}
\maketitle
\begin{abstract}

% High-quality data selection is crucial for the instruction fine-tuning of large language models (LLMs). In recent years, gradient-based data attribution methods have garnered widespread attention. However, this paper reveals the issue of surface form bias' inherent in traditional global gradient attribution methods: the gradients of model parameters are highly susceptible to formatting noise, leading to highly inconsistent training data attribution for instances sharing identical semantics. To overcome this challenge, we propose a novel framework named Content-Invariant Gradient Attribution (CIGA). This framework first introduces a dynamic layer selection algorithm to locate the gradients of the consistent influential layer, which are robust to formatting variations and possess high influence, thereby reducing surface-level noise within dynamic training features. Subsequently, a dynamic feature fusion mechanism is designed to integrate the gradient-based attribution signals with the underlying semantic features of the text, achieving a complement between the two. We conducted extensive experiments across multiple benchmarks, and the results demonstrate that CIGA improves the selection quality of training data while enhancing efficiency, providing an extremely robust and efficient new paradigm for data selection in LLMs training

Gradient-based data attribution has become an important tool for high-quality data selection in the instruction fine-tuning of large language models (LLMs), as it estimates the contribution of training examples through their influence on model optimization. However, we find that existing global gradient attribution methods suffer from surface-form bias: minor changes to semantically irrelevant details, such as wording or formatting, can produce highly inconsistent attribution scores even when the underlying content remains unchanged. To address this issue, we propose Content-Invariant Gradient Attribution (CIGA), a robust framework for instruction data selection. CIGA first identifies layers whose gradients are both influential for attribution and stable under semantics-preserving surface-form perturbations, thereby reducing formatting noise in gradient features. It then dynamically fuses these attribution signals with semantic representations of the input, combining optimization-level influence with content-level information. Extensive experiments across multiple benchmarks show that CIGA produces more stable rankings under surface-form perturbations while preserving downstream capabilities and improving average performance. CIGA also reduces attribution cost, offering a robust and efficient approach to data selection for LLM instruction tuning.

\end{abstract}

\section{Introduction}
\label{sec:intro}

\begin{figure}[t]
  \centering
  \includegraphics[width=\columnwidth]{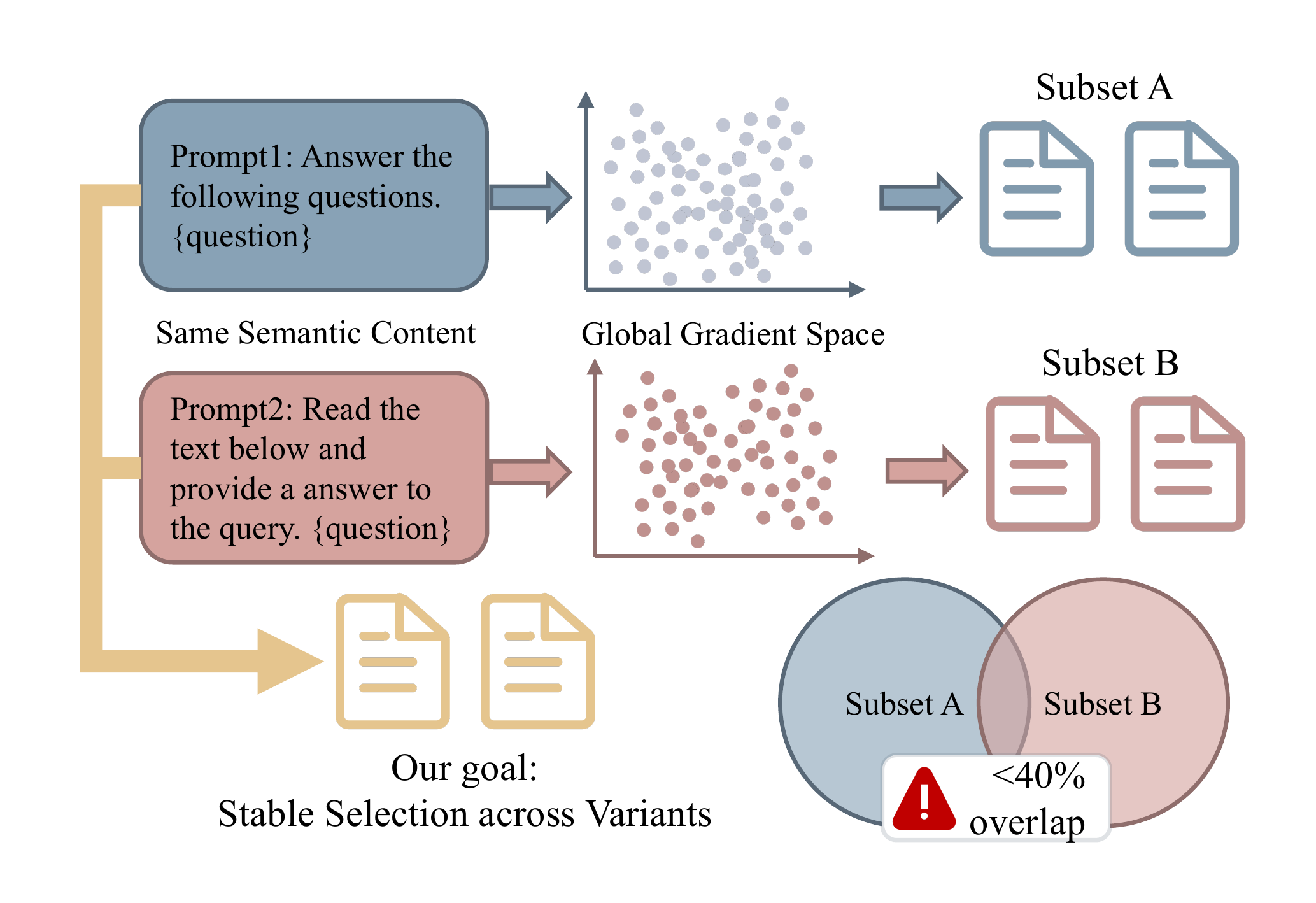}
  \caption{Global-gradient attribution can be sensitive to semantically irrelevant surface details: minor wording or formatting changes produce different gradient representations and low-overlap selected subsets.}
  \label{fig:motivation}
\end{figure}

Instruction tuning increasingly relies on large, automatically assembled corpora, which often contain redundant, noisy, or weakly informative examples. Recent work shows that models trained on carefully selected subsets can match or even outperform those trained on substantially larger datasets \citep{survey,Xia2024LESSSI}, highlighting target-aware data selection as a key component of efficient LLM post-training.

Targeted gradient-attribution methods rank training examples by the alignment between their parameter updates and those induced by target validation data \citep{Xia2024LESSSI}. This optimization-aware signal is useful, but global gradients also encode semantically irrelevant surface details such as wording and layout. As illustrated in Figure \ref{fig:motivation}, making minor surface-form changes to a validation example while preserving its content can change the selected training subset. We refer to this failure mode as \textit{surface-form bias}: attribution should respond primarily to the underlying task content rather than incidental forms of expression.

Semantic retrieval offers a complementary content-level signal \citep{GMValuator}, but similarity alone does not indicate whether a candidate induces a useful model update \citep{ImportanceAwareDS}. Robust targeted selection therefore requires both optimization-level relevance and reduced sensitivity to semantics-preserving surface variations.

\begin{figure*}[t]
  \centering
  \includegraphics[width=\linewidth]{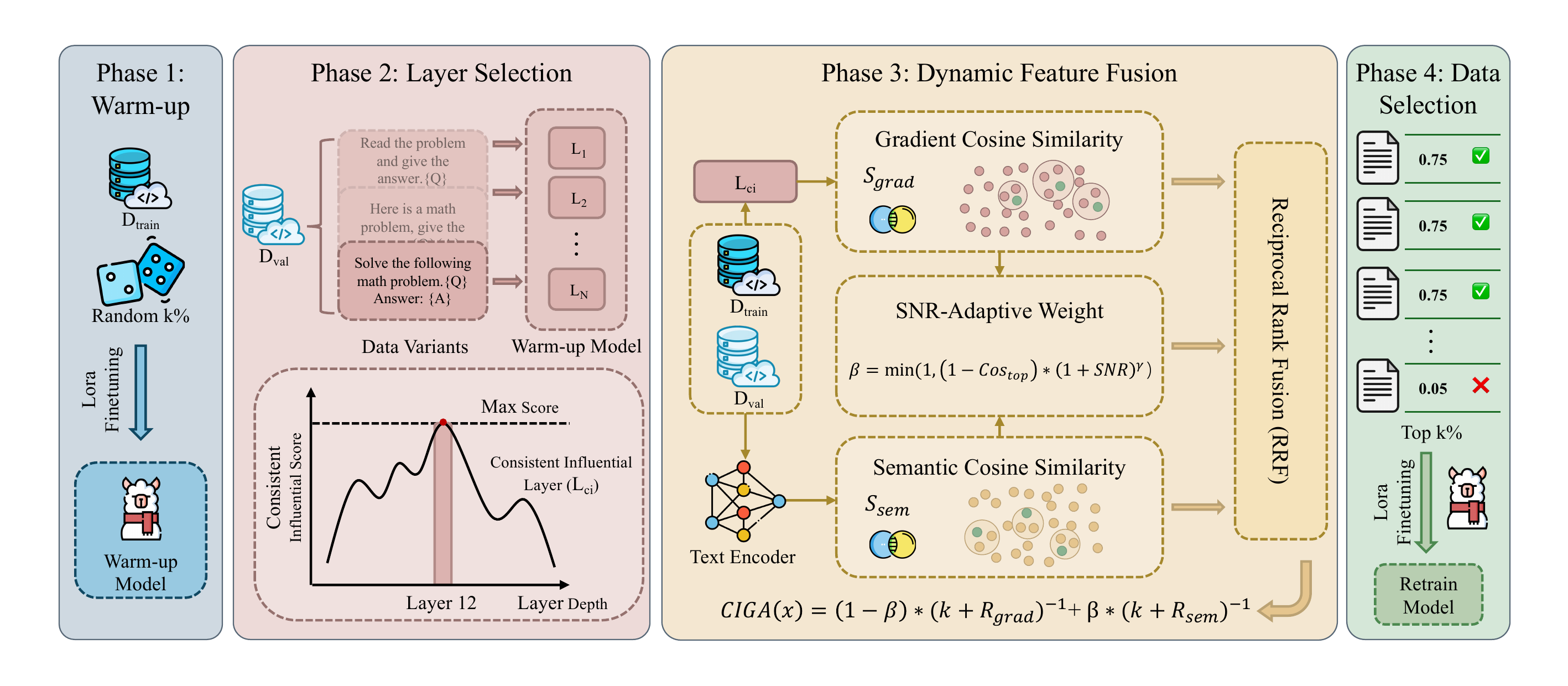}
\caption{Overview of CIGA: (1) warm-up fine-tuning; (2) consistent influential layer selection across prompt variants; (3) task-adaptive fusion of gradient and semantic rankings; and (4) Top-$k\%$ selection and retraining.}
  \label{fig:overview}
\end{figure*}

We propose \textbf{Content-Invariant Gradient Attribution (CIGA).} Motivated by evidence that Transformer layers encode different types of information \citep{Meng2022LocatingAE,Valeriani2023TheGO}, CIGA searches for a layer whose update features are strong yet stable across variants that differ only in semantically irrelevant surface form. It then computes train-validation gradient similarity only at this consistent influential layer. Finally, task-adaptive Reciprocal Rank Fusion (RRF) uses semantic similarity as a calibration signal rather than a replacement for gradient attribution. As summarized in Figure \ref{fig:overview}, the resulting pipeline uses one warm-up checkpoint and one selected layer, improving cross-variant consistency while preserving downstream utility. Importantly, CIGA imposes this invariance only during data attribution; it neither rewrites nor removes the original format diversity of the selected training data.

The main contributions are as follows:

    \textbf{(i) Characterizing surface-form bias in data attribution:} We show that minor, semantics-preserving changes to validation examples can produce substantially different global-gradient rankings and selected subsets.
    
    \textbf{(ii) Consistent influential layer selection:} 
    We introduce a layer score that balances update strength against cross-variant dispersion, yielding gradient features that are less sensitive to semantically irrelevant surface form.
    
    \textbf{(iii) Robust and effective data selection:} CIGA calibrates the selected-layer gradient ranking with semantic information. Across five benchmarks, it improves selection consistency while maintaining downstream capabilities and improving average performance, with lower attribution cost than multi-checkpoint gradient selection.

\section{Related Work}

\subsection{Data Attribution}

Training data attribution quantifies each training example's contribution to model predictions or performance. Classical Influence Functions (IF) \citep{grad,relatif} approximate the effect of removing a data point via the inverse Hessian, while recent work develops scalable approximations for LLMs \citep{scale-if,trak}. First-order methods such as TracIn \citep{tracin} and LESS \citep{Xia2024LESSSI} instead score training examples by their gradient similarity to validation examples, without the Hessian.

In addition to gradient-based approaches, another crucial direction is introducing Data Shapley based on cooperative game theory \citep{datashapley,scm} to guarantee theoretical fairness in data value allocation. In-Run Data Shapley \citep{in-Run-Data-Shapley} utilizes Taylor expansion within a single backward pass to achieve efficient approximate computation.

% In proactive attribution and copyright protection scenarios, watermark-based technologies exhibit unique advantages \citep{water2}. Such methods implicitly embed exclusive signals into the training corpus, prompting the model to learn and internalize these features during training, thereby providing a solution for data source verification. 

Regarding evaluation paradigms, the theoretical validity of data attribution algorithms typically requires rigorous verification through Leave-One-Out (LOO) retraining. However, due to the high computational cost of repeatedly retraining LLMs, current research also relies on utility evaluations based on downstream tasks. Prior work evaluates attribution scores in applications such as hallucination detection, fact tracing, and high-quality data selection \citep{Chen2024FASTTRACKFA, Wu2024EnhancingTD}.

\subsection{Data Selection}

For instruction tuning, data selection aims to identify a compact subset that maximizes performance on the target task. Gradient-based methods rank candidates using optimization signals. Recent work improves scalability through influence distillation \citep{Nikdan2025EfficientDS} and early-stage gradient clustering \citep{clusterucb}.
Beyond computational efficiency, inherent noise similarly constrains the performance of gradient methods. G-SNR \citep{Yuan2026UncertaintyAwareGS} uses gradient decay and variance during fine-tuning to estimate learning value and uncertainty, respectively. 
However, they generally rely only on global gradient features and struggle to capture the semantic relationships within the data, causing the selection results to lack robustness against variations in expression forms.

Complementary methods use semantic similarity, model-internal signals, or information-theoretic criteria for data selection \citep{GMValuator,Chen2026NeuronAwareDS,Duan2025EnhancingLV,Dong2026GreedyIP}. T-SHIRT \citep{Fu2025TSHIRTTH} exclusively focuses on highly informative key tokens within the response sequence and introduces neighborhood perturbations to filter out redundant samples that rely on spurious lexical features. Meanwhile, methods based on the ICL single-shot loss difference \citep{ImportanceAwareDS} have also been proven effective in quantifying the value of specific data in bridging model capability gaps.

Overall, global-gradient methods capture update utility but are sensitive to surface-form variations, whereas semantic and internal-signal methods capture content but are less tied to gradient dynamics.

\section{Methods}
\label{sec:methods}

The CIGA pipeline comprises warm-up fine-tuning, consistent influential layer selection, dynamic feature fusion, and subset selection followed by retraining. Its core components are the layer-selection score and task-adaptive fusion.

\subsection{Warm-up Fine-tuning}
\label{subsec:warmup}

Following LESS \citep{Xia2024LESSSI}, we randomly sample a subset $\mathcal{D}{warmup}\subset\mathcal{D}{train}$ and fine-tune the base model $M_{\theta}$, obtaining a warm-up checkpoint $M_{\theta^*}$ together with its optimizer state. We use this state to compute Adam-preconditioned update features, and perform both layer selection and attribution from this single checkpoint.

\subsection{Consistent Influential Layer Selection}
\label{subsec:layer_selection}

Prior work has shown that different Transformer layers capture distinct types of information \citep{Meng2022LocatingAE,Valeriani2023TheGO}. Building on this observation, we identify the layer whose update features exhibit high average magnitude while remaining stable across semantics-preserving surface-form variants.

For each example $x_i \in \mathcal{D}_{val}$, we construct $m$ variants $\{p_j(x_i)\}_{j=1}^m$ through minor changes to semantically irrelevant wording or formatting while preserving the underlying content. $\Gamma_{i,l}^{(j)}$ denotes the Adam-preconditioned update feature at layer $l$. We characterize each layer by its average update magnitude $\mu_l$ (update strength) and the dispersion of its update vectors across variants $\sigma_l^2$ (instability):
\begin{equation}
\begin{aligned}
\mu_{l} &= \mathbb{E}_{i}\,\mathbb{E}_{j} \big\|\Gamma_{i,l}^{(j)}\big\|_2 , \\
\sigma_{l}^{2} &= \mathbb{E}_{i}\,\mathbb{E}_{j} \big\|\Gamma_{i,l}^{(j)} - \bar{\Gamma}_{i,l}\big\|_2^2 ,
\end{aligned}
\end{equation}
where $\bar{\Gamma}_{i,l}=\mathbb{E}_{j}[\Gamma_{i,l}^{(j)}]$. Thus, $\sigma_l$ captures cross-variant variation in both the magnitude and direction of the update vectors.

We Z-score $\mu_l$ and $\sigma_l$ across layers to obtain $\hat{\mu}_l$ and $\hat{\sigma}_l$. The consistent influential score is
\begin{equation}
S_{l}^{ci} = \hat{\mu}_{l} - \alpha \cdot \hat{\sigma}_{l}
\end{equation}
where $\alpha$ controls the penalty on cross-variant instability. We select layer $l_{ci}=\arg\max_l S_l^{ci}$.

\subsection{Dynamic Feature Fusion}
\label{subsec:feature_fusion}

After selecting $l_{ci}$, we score each candidate $x\in\mathcal{D}_{train}$ using gradient-based relevance and semantic similarity to the target validation set.

\textbf{Grad-Score:} We extract the gradients from $l_{ci}$, flattening and concatenating them into a unified gradient vector. For efficient storage, dimensionality reduction is applied in practical implementations to obtain a gradient vector $G$. We then calculate the cosine similarity, denoted as $S_{grad}$, to measure the consistency between the training data and the validation data in the direction of parameter updates.

\textbf{Sem-Score:} Using a external text embedding model, we extract the text representation $E$ of the data. We then calculate the cosine similarity in the semantic representation space to reflect the surface expression similarity of the data, yielding $S_{sem}$.

Because gradient and semantic similarities have misaligned distributions, direct addition is unstable. Thus, we compute the final CIGA score via RRF. Let $R_{grad}(x)$ and $R_{sem}(x)$ be the rankings of the candidate data based on the Grad-Score and Sem-Score, respectively. The CIGA score is:

\begin{equation}
\label{eq:rrf}
\small
\text{CIGA}(x) = (1 - \beta) \cdot \frac{1}{r + R_{grad}(x)} + \beta \cdot \frac{1}{r + R_{sem}(x)}
\end{equation}
where $r$ is a smoothing constant. The dynamic weight $\beta \in [0, 1]$ controls the degree of reliance on semantic features during fusion.

\paragraph{Dynamic Fusion Ratio:}
We derive a task-specific $\beta$ from two diagnostics rather than assigning a uniform weight to all tasks.

\textit{How reliable is the gradient view?} Let $\Gamma_{i}^{(j)} \equiv \Gamma_{i,l_{ci}}^{(j)}$. Since the $m$ variants preserve the content of each sample, within-sample spread reflects semantically irrelevant surface-form noise, while between-sample spread reflects distinctions that attribution should preserve. This motivates an ANOVA-style signal-to-noise decomposition \citep{fisher1925statistical}:
\begin{equation}
\mathrm{SNR} = \frac{\text{Var}_{i}\big(\mathbb{E}_{j}[\Gamma_{i}^{(j)}]\big)}{\mathbb{E}_{i}\big[\text{Var}_{j}(\Gamma_{i}^{(j)})\big] + \epsilon}
\end{equation}
where variance of a vector denotes coordinate-wise variance averaged over its dimensions. 
With high $\mathrm{SNR}$, examples remain distinguishable despite surface-form perturbations, so $\beta$ should shrink.

\textit{How concentrated is the semantic view?} A low $\mathrm{SNR}$ favors greater semantic support. However, when the highest Sem-Scores are already very large, relying too heavily on the semantic ranking concentrates selection on candidates that are merely more similar to the validation data. The resulting subset covers a narrower semantic neighborhood and therefore has lower diversity. We quantify this tendency using $\mathrm{Sem}^{Top}$, the mean of the top $K$ aggregated per-candidate Sem-Scores; a larger value reduces $\beta$ to preserve a better balance between semantic relevance and data diversity.

Combining both diagnostics gives the fusion weight for the validation set:
\begin{equation}
\beta = \min\left( 1,\ \frac{1 - \mathrm{Sem}^{Top}}{(1 + \mathrm{SNR})^\gamma} \right)
\end{equation}
where $\gamma$ controls the SNR sensitivity and $1+\mathrm{SNR}$ keeps the denominator stable at small SNR. Both diagnostics are properties of the validation set, so each task gets one $\beta$, while Eq.~\eqref{eq:rrf} acts on each candidate through its own ranks: task-level diagnosis and sample-level selection stay decoupled.

\subsection{Data Selection and Retraining}
\label{subsec:retraining}

For a target validation set, we rank all candidates in $\mathcal{D}_{train}$ by their CIGA scores and select the Top-$k\%$ subset. We then retrain the original base model $M_{\theta}$ on this subset and evaluate its target-task performance.

\section{Experiment}
% \subsection{Setup}
% train: "flan_v2", "cot", "dolly", "oasst1", "alpaca_gpt4", "gsm8k_train"
% test: tydiqa, bbh, gsm8k, mmlu, hellaswag, arc-c
% model: llama2-7b,qwen3-8b,qwen3-4b
% 评估：性能，鲁棒性（重合度）
% base：cos，bm25，less，ifd，t-shirt

\subsection{Experimental Setup}

% \textbf{Datasets.} 
% To comprehensively evaluate our proposed CIGA framework, we construct a diverse training dataset from several widely adopted instruction tuning collections. These comprise FLAN-v2 \citep{flanv2}, Chain of Thought (CoT) \citep{cot}, Dolly \citep{dolly}, OpenAssistant (OASST1) \citep{oasst}, Alpaca-GPT4 \citep{alpaca}, and the GSM8K training set \citep{gsm8k}, ensuring a rich mixture of general instructions, mathematical reasoning, and multi-turn dialogues.

% For downstream evaluation, we select five rigorous benchmarks spanning various capabilities: TyDiQA (multilingual understanding) \citep{tydiqa}, BBH (complex algorithmic and logical reasoning) \citep{bbh}, GSM8K (mathematical problem solving) \citep{gsm8k}, MMLU (massive multitask language understanding) \citep{mmlu}, and HellaSwag (commonsense reasoning) \citep{hellaswag}. 

% \textbf{Models.}
% We validate the generalizability and scalability of our data selection methods across different foundation models. Specifically, we employ LLaMA-2-7B \citep{llama2}, alongside the Qwen3-4B-Base \citep{qwen3}, serving as the base architectures for both the data attribution phase and the subsequent instruction fine-tuning.

\textbf{Datasets and Models.} We construct a diverse instruction-tuning pool from FLAN-v2 \citep{flanv2}, CoT \citep{cot}, Dolly \citep{dolly}, OASST1 \citep{oasst}, Alpaca-GPT4 \citep{alpaca}, and GSM8K \citep{gsm8k}. We evaluate on five benchmarks: TyDiQA \citep{tydiqa}, BBH \citep{bbh}, GSM8K \citep{gsm8k}, MMLU \citep{mmlu}, and HellaSwag \citep{hellaswag}. Detailed dataset statistics are provided in Appendix \ref{sec:appendix_datasets}.
We use LLaMA-2-7B \citep{llama2} and Qwen3-4B-Base \citep{qwen3} for attribution and fine-tuning

% \textbf{Models.} We employ LLaMA-2-7B \citep{llama2} and Qwen3-4B-Base \citep{qwen3} for both data attribution and subsequent instruction fine-tuning.

% \textbf{Baselines.}
% We compare CIGA against a comprehensive set of strong data selection baselines, which can be broadly categorized into three groups:

% \textit{Semantic Matching Methods:} cosine similarity (using all-MiniLM-L6-v2 \citep{reimers-2019-sentence-bert} for dense retrieval) to retrieve training samples most similar to the validation set texts.  

% \textit{Gradient-based Attribution:} LESS \citep{Xia2024LESSSI}, TracIn \citep{tracin}, selecting data based on parameter influence.  

% \textit{Information based Methods:} IFD \citep{ifd}, T-SHIRT \citep{Fu2025TSHIRTTH}, and SelectIT \citep{liu2024selectit}, selecting data based on instruction following difficulty and model confidence. Given our massive training set scale, we reproduce the SelectIT model level score only for LLaMA-2-7B and LLaMA-2-13B, consistent with the original paper.

\textbf{Baselines.} We compare CIGA with (1) \textit{Semantic Matching}: cosine similarity via dense retrieval \citep{reimers-2019-sentence-bert}; (2) \textit{Gradient Methods}: LESS \citep{Xia2024LESSSI} and TracIn \citep{tracin}; and (3) \textit{Information-based Methods}: IFD \citep{ifd} and SelectIT \citep{liu2024selectit}. Details are provided in Appendix \ref{sec:appendix_baselines}.

% \textbf{Evaluation Metrics.}
% We assess the selected data subsets from two primary perspectives:

% \textit{Downstream Performance:} We report benchmark specific metrics (e.g., accuracy or exact match) after retraining the base models exclusively on the top $k$\% selected data subsets.

% \textit{Robustness:} To explicitly quantify the surface form bias, we generate multiple semantic-equivalent prompt variants and compute the overlap rate (Jaccard similarity \citep{jaccard1912distribution}) among the respective Top-$k$\% subsets for the same validation data. A higher overlap indicates that the selection algorithm is genuinely driven by content rather than being disturbed by surface formats.

% \textbf{Evaluation Metrics.} We assess selected subsets from two perspectives. \textit{Selection consistency} is the Jaccard similarity \citep{jaccard1912distribution} between Top-$k\%$ subsets obtained from different semantics-preserving surface-form variants. With five variants, we average all $\binom{5}{2}=10$ pairwise Jaccard values. \textit{Downstream performance} is measured by TyDiQA F1, MMLU and HellaSwag accuracy, and BBH and GSM8K exact match after retraining on the selected subset. 

\textbf{Evaluation.} We measure selection consistency by mean pairwise Jaccard similarity \citep{jaccard1912distribution} among Top-$k\%$ subsets from semantics-preserving surface-form variants, and downstream performance by task metrics after retraining. Protocol and metric details are provided in Appendix \ref{app:templates}.

\textbf{Settings.} The warm-up and selected subsets each contain 5\% of the candidate pool. We choose $\alpha=0.7$ and $\gamma=1/3$ on TyDiQA and freeze them across the remaining tasks and both backbones. We use LoRA \citep{lora}; remaining hyperparameters, training settings, and hardware details are provided in Appendix \ref{app:training_parameters}, with hyperparameters sensitivity results in Appendix \ref{app:hyperparameter}.

% \textbf{Surface-Form Variant Protocol.} For each target task, we manually construct five variants of the same validation examples (Appendix \ref{app:templates}). They preserve the original task semantics and question/answer content while making only minor changes to semantically irrelevant wording or layout. CIGA, LESS, and TracIn use the same five variants in the consistency evaluation.
% % Cosine retrieval encodes the question and answer without the prompt wrapper, so it is prompt-agnostic and does not enter the Jaccard comparison.

% \textbf{Hyperparameter Settings.} 
% The warm-up subset and final selected subset each contain 5\% of the candidate pool. We set the RRF smoothing constant $r=60$, the semantic-head size $K=1000$, and $\epsilon=10^{-8}$. We choose $\alpha=0.7$ and $\gamma=1/3$ on TyDiQA and then freeze this configuration for the remaining test sets and both backbones; Appendix \ref{app:hyperparameter} reports sensitivity.

% \textbf{Implementation Details.} 
% We use Low-Rank Adaptation (LoRA) \citep{lora} for parameter updates (Appendix \ref{app: traing-para}). CIGA performs attribution from one warm-up checkpoint, whereas LESS aggregates attribution gradients from four checkpoints. After selecting one subset, each method retrains the base model once per run. All core experiments were conducted on an Ubuntu Linux server with 4 NVIDIA A100 GPUs (80GB).

\subsection{Experimental Result}
\subsubsection{Selection Consistency Across Variants}

To evaluate the sensitivity of gradient-based data selection to semantically irrelevant surface-form variations, we perform data selection using five variants of the same validation data and compute the mean pairwise Jaccard similarity among the resulting Top-5\% training subsets.

\textbf{Overall Advantage over Global-Gradient Baselines.} As shown in Table \ref{tab:overlap_results}, CIGA achieves average overlap rates of 58.33\% and 48.68\% on LLaMA-2-7B and Qwen3-4B-Base, respectively. Compared with the best-performing global-gradient baseline on each backbone, these results represent improvements of 9.27 and 8.65 percentage points. The gains span diverse tasks, including MMLU, BBH, GSM8K, and HellaSwag, showing that CIGA produces more consistent selections across model architectures and tasks.
    
\textbf{Task-Level Analysis of Surface-Form Sensitivity.} HellaSwag further illustrates the effect of surface form on gradient attribution. When the underlying context and answer content are preserved while only instruction wording are changed, CIGA achieves overlap rates of 49.29\% and 36.60\%, approximately twice the corresponding LESS values (22.64\% and 18.47\%). Because these variants share the same underlying content, the increased overlap directly reflects lower sensitivity to surface expression, showing that CIGA effectively mitigates surface-form bias in global-gradient attribution.

\begin{table}[t]
  \centering
  \caption{Mean pairwise Jaccard similarity (\%) across five semantics-preserving surface-form variants. Higher values indicate greater selection consistency.}

  \label{tab:overlap_results}
  \begin{tabular}{l | ccc}
    \toprule
    \textbf{Test} & \textbf{LESS} & \textbf{TracIn} & \textbf{CIGA} \\
    \midrule
    \multicolumn{4}{c}{\textbf{LLaMA-2-7B}} \\
    \midrule
    TyDiQA      & 41.40 & 36.75 & \textbf{42.40} \\
    MMLU        & 62.16 & 58.50 & \textbf{76.04} \\
    BBH         & 68.94 & 69.86 & \textbf{71.92} \\
    GSM8K       & 46.07 & 51.96 & \textbf{52.00} \\
    HellaSwag   & 22.64 & 28.24 & \textbf{49.29} \\
    \midrule
    \textbf{Avg} & 48.24 & 49.06 & \textbf{58.33} \\
    \midrule
    \multicolumn{4}{c}{\textbf{Qwen3-4B-Base}} \\
    \midrule
    TyDiQA      & \textbf{62.75} & 37.29 & 50.03 \\
    MMLU        & 57.03 & 55.49 & \textbf{68.06} \\
    BBH         & 48.62 & 50.59 & \textbf{67.11} \\
    GSM8K       & 13.28 & 17.77 & \textbf{21.60} \\
    HellaSwag   & 18.47 & 17.25 & \textbf{36.60} \\
    \midrule
    \textbf{Avg} & 40.03 & 35.68 & \textbf{48.68} \\
    \bottomrule
  \end{tabular}
\end{table}

\subsubsection{Downstream Performance Results}

To evaluate the downstream utility of the selected data, we fine-tune the base models on the Top-5\% training subsets selected by different methods and evaluate them on five target tasks. In Table \ref{tab:performance_comparison}, ``Full'' denotes fine-tuning on the complete dataset.

\textbf{Overall Performance.} CIGA achieves the highest macro-average among all data selection methods on both backbones, exceeding Random by 2.11 and 1.04 percentage points on LLaMA-2-7B and Qwen3-4B-Base, respectively. Paired tests on the run-level macro-averages show significant gains against all evaluated selection baselines on LLaMA-2-7B and against Random, Cos, and SelectIT on Qwen3-4B-Base ($p<0.05$). Under the challenging Top-5\% setting, these gains are comparable to those reported in recent data selection studies \citep{clusterucb,Chen2026NeuronAwareDS,ImportanceAwareDS}, while the consistency results above show that CIGA improves robustness without sacrificing downstream utility.

\textbf{Complementarity of Semantic and Gradient Signals.} The strong performance of Cos on GSM8K demonstrates the value of semantic similarity for mathematical word problems with prominent structural and semantic patterns. By dynamically combining semantic similarity with gradient attribution, CIGA outperforms global-gradient methods on GSM8K for both backbones. Meanwhile, CIGA maintains more balanced performance on the other tasks, indicating that semantic and gradient signals can effectively complement each other according to the characteristics of the target task.

\textbf{Approaching Full-Data Fine-Tuning.} Using only 5\% of the training data, data selection methods approach full-data fine-tuning on multiple tasks. For Qwen3-4B-Base, the macro-average of Full is 1.12 percentage points lower than that of Random, suggesting that more capable base models may be more sensitive to low-value or noisy fine-tuning data and that indiscriminately using the complete candidate pool may weaken their original capabilities. By filtering out lower-value samples, data selection methods achieve higher averages than Full, with CIGA attaining the best overall performance.

On LLaMA-2-7B, Full retains a slightly higher average than CIGA, but CIGA surpasses Full on TyDiQA and MMLU. These results show that CIGA maintains downstream performance using a smaller training subset while achieving an overall improvement over Random, demonstrating data efficiency.

\begin{table*}[htbp]
  \centering
  \caption{Performance comparison across test datasets (\%). $\Delta_{Rand}$ is the average improvement over Random, and parentheses report standard deviations over three runs. $*$ marks a significant improvement of CIGA over Random ($p<0.05$) on the corresponding task or macro-average. \textbf{Bold} marks the best result among selection methods (excluding Full and Random), and \underline{underline} marks performance surpassing Full.
}
  \label{tab:performance_comparison}
\resizebox{\textwidth}{!}{
  \begin{tabular}{l | cccccc | cccccc}
    \toprule
    \multirow{2}{*}{\textbf{Method}} & \multicolumn{6}{c|}{\textbf{LLaMA-2-7B}} & \multicolumn{6}{c}{\textbf{Qwen3-4B-Base}} \\
    \cmidrule{2-13}
    & \textbf{TyDiQA} & \textbf{MMLU} & \textbf{BBH} & \textbf{GSM} & \textbf{Hella} & \textbf{$\Delta_{Rand}$} & \textbf{TyDiQA} & \textbf{MMLU} & \textbf{BBH} & \textbf{GSM} & \textbf{Hella} & \textbf{$\Delta_{Rand}$} \\
    \midrule
    \multicolumn{13}{l}{\textit{\textcolor{gray}{Reference Baselines}}} \\
    Full            & 50.77 {\scriptsize (0.3)} & 47.14 {\scriptsize (0.2)} & 43.80 {\scriptsize (0.8)} & 25.17 {\scriptsize (1.0)} & 74.15 {\scriptsize (0.3)} & \textcolor{green!60!black}{+2.48} & 66.34 {\scriptsize (0.4)} & 72.89 {\scriptsize (0.3)} & 68.52 {\scriptsize (0.9)} & 78.09 {\scriptsize (1.1)} & 70.40 {\scriptsize (0.3)} & \textcolor{red}{-1.12} \\
    Random          & 49.54 {\scriptsize (0.5)} & 45.51 {\scriptsize (0.4)} & 41.85 {\scriptsize (0.6)} & 17.29 {\scriptsize (1.6)} & \underline{74.47} {\scriptsize (0.4)} & - & \underline{73.16} {\scriptsize (0.5)} & \underline{73.01} {\scriptsize (0.4)} & 67.13 {\scriptsize (1.3)} & 78.01 {\scriptsize (0.9)} & \underline{70.52} {\scriptsize (0.4)} & - \\
    \midrule
    \multicolumn{13}{l}{\textit{\textcolor{gray}{Heuristic \& Semantic Methods}}} \\
    Cos             & 48.31 {\scriptsize (0.4)} & 46.06 {\scriptsize (0.3)} & 40.93 {\scriptsize (1.0)} & \textbf{21.91} {\scriptsize (0.7)} & \underline{74.52} {\scriptsize (0.3)} & \textcolor{green!60!black}{+0.62} & \underline{72.23} {\scriptsize (1.4)} & \underline{73.23} {\scriptsize (0.3)} & 67.41 {\scriptsize (1.1)} & \underline{78.85} {\scriptsize (1.0)} & \underline{70.69} {\scriptsize (0.3)} & \textcolor{green!60!black}{\underline{+0.11}} \\
    IFD             & 43.21 {\scriptsize (0.6)} & 44.45 {\scriptsize (0.5)} & \textbf{41.67} {\scriptsize (0.7)} & 14.78 {\scriptsize (0.5)} & \underline{74.17} {\scriptsize (0.4)} & \textcolor{red}{-2.08} & \underline{71.58} {\scriptsize (0.5)} & 72.86 {\scriptsize (0.4)} & 67.87 {\scriptsize (1.0)} & \underline{79.15} {\scriptsize (1.2)} & \underline{70.44} {\scriptsize (0.4)} & \textcolor{green!60!black}{\underline{+0.01}} \\
    SelectIT        & 49.96 {\scriptsize (0.7)} & 45.91 {\scriptsize (0.4)} & \textbf{41.67} {\scriptsize (0.5)} & 20.39 {\scriptsize (0.6)} & \underline{74.35} {\scriptsize (0.3)} & \textcolor{green!60!black}{+0.72} & \underline{71.37} {\scriptsize (0.5)} & \underline{73.21} {\scriptsize (0.3)} & 67.87 {\scriptsize (0.6)} & 77.03 {\scriptsize (1.3)} & \underline{\textbf{70.73}} {\scriptsize (0.3)} & \textcolor{red}{\underline{-0.33}} \\
    \midrule
    \multicolumn{13}{l}{\textit{\textcolor{gray}{Gradient-based Methods}}} \\
    LESS            & \underline{54.65} {\scriptsize (0.5)} & 45.19 {\scriptsize (0.8)} & 40.00 {\scriptsize (1.1)} & 17.59 {\scriptsize (1.2)} & \underline{74.50} {\scriptsize (0.4)} & \textcolor{green!60!black}{+0.66} & \underline{73.38} {\scriptsize (1.6)} & \underline{\textbf{73.37}} {\scriptsize (0.4)} & \underline{69.63} {\scriptsize (0.5)} & \underline{78.17} {\scriptsize (1.6)} & 70.36 {\scriptsize (0.4)} & \textcolor{green!60!black}{\underline{+0.61}} \\
    TracIn          & \underline{54.05} {\scriptsize (1.2)} & 45.52 {\scriptsize (0.5)} & 40.74 {\scriptsize (1.2)} & 16.76 {\scriptsize (1.1)} & 73.71 {\scriptsize (0.6)} & \textcolor{green!60!black}{+0.43} & \underline{72.05} {\scriptsize (0.6)} & 72.67 {\scriptsize (0.5)} & 68.33 {\scriptsize (1.5)} & \underline{78.32} {\scriptsize (0.8)} & 69.44 {\scriptsize (0.5)} & \textcolor{red}{\underline{-0.21}} \\
    \midrule
    \multicolumn{13}{l}{\textit{\textcolor{gray}{Ours}}} \\
    \textbf{CIGA}   & \underline{\textbf{56.17$^{*}$}} {\scriptsize (0.2)} & \underline{\textbf{47.46$^{*}$}} {\scriptsize (0.2)} & 41.39 {\scriptsize (0.6)} & 19.64$^{*}$ {\scriptsize (1.0)} & \underline{\textbf{74.54}} {\scriptsize (0.2)} & \textbf{\textcolor{green!60!black}{+2.11$^{*}$}} & \underline{\textbf{74.10$^{*}$}} {\scriptsize (0.2)} & \underline{73.05} {\scriptsize (0.2)} & \underline{\textbf{70.28$^{*}$}} {\scriptsize (0.7)} & \underline{\textbf{79.30$^{*}$}} {\scriptsize (1.2)} & 70.29 {\scriptsize (0.2)} & \textbf{\textcolor{green!60!black}{\underline{+1.04$^{*}$}}} \\
    \bottomrule
  \end{tabular}
  }
\end{table*}

\subsubsection{Ablation Experiment}
\textbf{Consistent Influential Layer Selection.}
To evaluate the contribution of the layer selection module, we construct CIGA-G by removing the Sem-Score and selecting data using only the Grad-Score extracted from the consistent influential layer.

As shown in Table \ref{tab:component_ablation}, CIGA-G achieves average Jaccard similarities of 56.44\% and 46.96\% on LLaMA-2-7B and Qwen3-4B-Base, respectively, with downstream averages of 47.15\% and 73.26\%; all four results exceed those of LESS and TracIn. These results show that extracting gradient features only from the consistent influential layer already provides most of the selection-stability gains while preserving performance. Compared with aggregating gradients across all layers, focusing on a more stable and task-relevant intermediate layer helps reduce the interference of surface-form signals unrelated to the underlying content.

\begin{table}[t]
  \centering
  \small
  \caption{Component ablation in selection consistency and downstream performance (\%).}
  \label{tab:component_ablation}
  \resizebox{\columnwidth}{!}{
  \begin{tabular}{llcc}
    \toprule
    \textbf{Model} & \textbf{Method} & \textbf{Jaccard} & \textbf{Downstream Avg} \\
    \midrule
    \multirow{2}{*}{LLaMA-2-7B}
      & CIGA-G & 56.44 & 47.15 \\
      & CIGA   & \textbf{58.33} & \textbf{47.84} \\
    \midrule
    \multirow{2}{*}{Qwen3-4B-Base}
      & CIGA-G & 46.96 & 73.26 \\
      & CIGA   & \textbf{48.68} & \textbf{73.40} \\
    \bottomrule
  \end{tabular}
  }
\end{table}

\textbf{Semantic and Gradient Feature Fusion.}
After adding dynamic semantic fusion to CIGA-G, full CIGA improves the average Jaccard similarity by 1.89 and 1.72 percentage points on LLaMA-2-7B and Qwen3-4B-Base, respectively, while maintaining and slightly improving the downstream performance. This shows that the Grad-Score from the consistent influential layer provides the main foundation for stable selection, while semantic information calibrates the gradient-only ranking to balance gradient-based utility with content relevance.

\begin{figure*}[t]
  \centering
  \includegraphics[width=\linewidth]{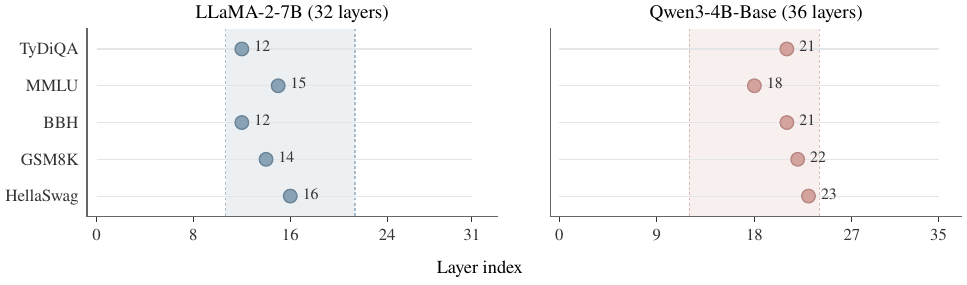}
  \caption{Selected consistent influential layers  across tasks. Shaded regions denote the middle layers of each model.}
  \label{fig:selected_layers}
\end{figure*}

\textbf{Layer Choice.}
As shown in Figure \ref{fig:selected_layers}, the layers selected by CIGA across different tasks consistently lie in the middle regions of both models. The selected layers range from 12 to 16 in LLaMA-2-7B (32 layers) and from 18 to 23 in Qwen3-4B-Base (36 layers). Across different random seeds, Qwen3 selects the same layer for every task, while LLaMA-2 exhibits only a small variation on BBH. These results show that CIGA does not depend on a fixed absolute layer index, but can stably locate a middle-layer region that combines gradient strength with surface-form stability for each task.

To examine whether aggregating multiple layers improves attribution, we rank layers by $S_l^{ci}$, select the top $k\in\{1,3,5\}$ layers, and average their layer-wise gradient similarity scores to obtain the Grad-Score. As shown in Table \ref{tab:layer_topk}, multi-layer aggregation produces mixed task-level effects. The best multi-layer setting improves over the single-layer setting by only 0.19 and 0.22 percentage points on LLaMA-2-7B and Qwen3-4B-Base, respectively. 
% Because using more layers increases gradient computation and storage costs, we retain a single consistent influential layer as the default setting.

\begin{table*}[t]
  \centering
  \small
  \caption{Effect of the number of attribution layers on downstream performance (\%). Layers are ranked by $S_l^{ci}$. Values are differences from the single-layer setting ($k=1$) measured in the same runs.}
  \label{tab:layer_topk}
  \begin{tabular}{llcccccc}
    \toprule
    \textbf{Model} & \textbf{Layers ($k$)} & \textbf{TyDiQA} & \textbf{MMLU} & \textbf{BBH} & \textbf{GSM} & \textbf{Hella} & \textbf{Avg} \\
    \midrule
    \multirow{2}{*}{LLaMA-2-7B}
      & 3 & +0.10 & $-$0.01 & $-$0.19 & +0.23 & $-$0.17 & $-$0.01 \\
      & 5 & +0.67 & +0.06 & +0.18 & +0.15 & $-$0.11 & +0.19 \\
    \midrule
    \multirow{2}{*}{Qwen3-4B-Base}
      & 3 & +0.67 & +0.12 & 0.00 & $-$0.22 & $-$0.10 & +0.09 \\
      & 5 & +1.04 & +0.08 & $-$0.09 & +0.08 & $-$0.02 & +0.22 \\
    \bottomrule
  \end{tabular}
\end{table*}

\textbf{Number of Surface-Form Variants.}
Surface-form variants are used to estimate gradient stability for the same content under different expressions, so the number of variants may affect the selected layer. As shown in Table \ref{tab:n_variants}, when the number of variants increases from three to five, the downstream average fluctuates by less than 0.4 percentage points on both models, indicating stable downstream performance under this setting. Meanwhile, the average Jaccard similarity increases by 3.09 and 1.96 percentage points on Qwen3 and LLaMA-2, respectively, showing that additional semantics-preserving variants provide a more comprehensive estimate of stability across surface forms.

\begin{table}[t]
  \centering
  \small
  \setlength{\tabcolsep}{4pt}
  \caption{Number of surface-form variants (\%). Values are differences from the default setting ($n=5$).}
  \label{tab:n_variants}
  \begin{tabular}{lcccc}
    \toprule
    \textbf{Model} & \multicolumn{2}{c}{\textbf{Downstream Avg}} & \multicolumn{2}{c}{\textbf{Jaccard}} \\
    \cmidrule(lr){2-3} \cmidrule(lr){4-5}
    & $n=3$ & $n=4$ & $n=3$ & $n=4$ \\
    \midrule
    Qwen3-4B-Base & $-$0.37 & $-$0.19 & $-$3.09 & $-$0.95 \\
    LLaMA-2-7B & $-$0.09 & $-$0.35 & $-$1.96 & $-$0.08 \\
    \bottomrule
  \end{tabular}
\end{table}

\subsubsection{Data Selection Efficiency}
Table \ref{tab:cost} reports the computational cost of each stage in the data-selection pipeline, with runtime normalized to GPU hours on a single A100 (80GB). Here, $N$ denotes the number of checkpoints used for attribution. Warm-up training and candidate-data gradient computation are one-time costs; when the model and candidate pool remain unchanged, their outputs can be reused across target tasks. CIGA uses single-layer gradients from one checkpoint, whereas LESS and TracIn aggregate global gradients from the same four checkpoints and thus have the same cost. Although CIGA additionally requires 5 minutes for layer selection and 6 minutes for Sem-Score computation, it reduces the total runtime from 74.1 to 21.4 hours, a reduction of approximately 71.1\% and a $3.46\times$ speedup.

\begin{table}[t]
  \centering
  \small
  \setlength{\tabcolsep}{4pt}
  \caption{Computational cost of the data-selection pipeline. $L$ and $m$ denote the numbers of layers and surface-form variants; $d_g$ and $d_s$ denote the gradient and semantic feature dimensions.}
  \label{tab:cost}
  \resizebox{\columnwidth}{!}{
  \begin{tabular}{llcc}
    \toprule
    \textbf{Stage} & \textbf{Complexity} & \textbf{CIGA} & \textbf{LESS} \\
    \midrule
    Warm-up
      & $\mathcal{O}(|\mathcal{D}_{\mathrm{warmup}}|\cdot N)$
      & 1 Hour & 4 Hours \\
    Grad-Score
      & $\mathcal{O}(|\mathcal{D}|\cdot N)$
      & 20 Hours & 70 Hours \\
    Layer Selection
      & $\mathcal{O}(mL|\mathcal{D}_{\mathrm{val}}|d_g)$
      & 5 Min & -- \\
    Sem-Score
      & $\mathcal{O}(|\mathcal{D}||\mathcal{D}_{\mathrm{val}}|d_s)$
      & 6 Min & -- \\
    Data Selection
      & $\mathcal{O}(|\mathcal{D}||\mathcal{D}_{\mathrm{val}}|d_g)$
      & $<$1 Min & $<$1 Min \\
    \midrule
    Total & & \textbf{21.4 Hours} & 74.1 Hours \\
    \bottomrule
  \end{tabular}
  }
\end{table}

\section{Discussion}

\subsection{Overcoming Surface Form Bias}
% 可视化同一条数据的不同prompt在梯度空间中的分散
% todo 在中间层+语义层的聚拢（待定）
\begin{figure*}[t]
  \includegraphics[width=0.33\linewidth]{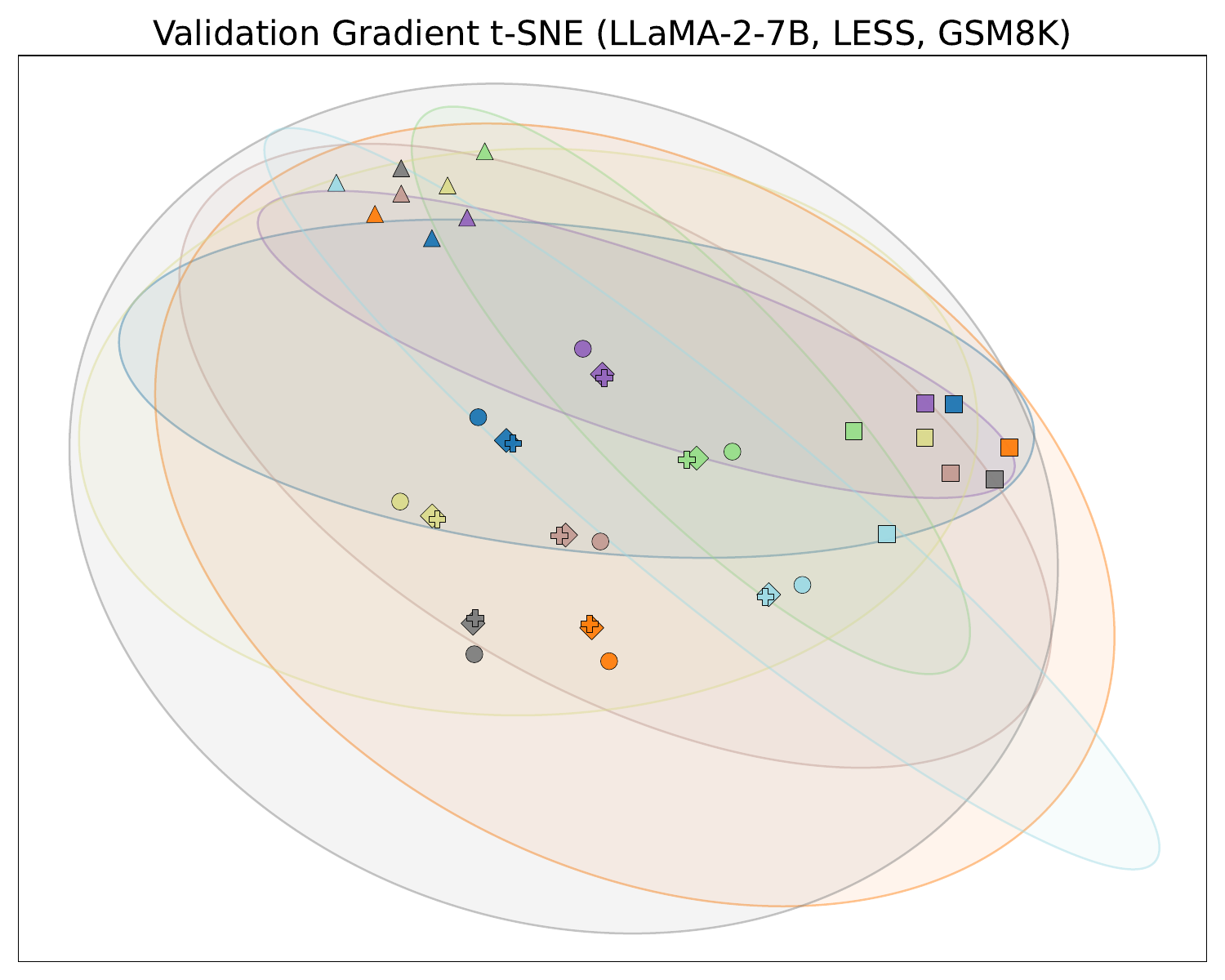}
  \includegraphics[width=0.33\linewidth]{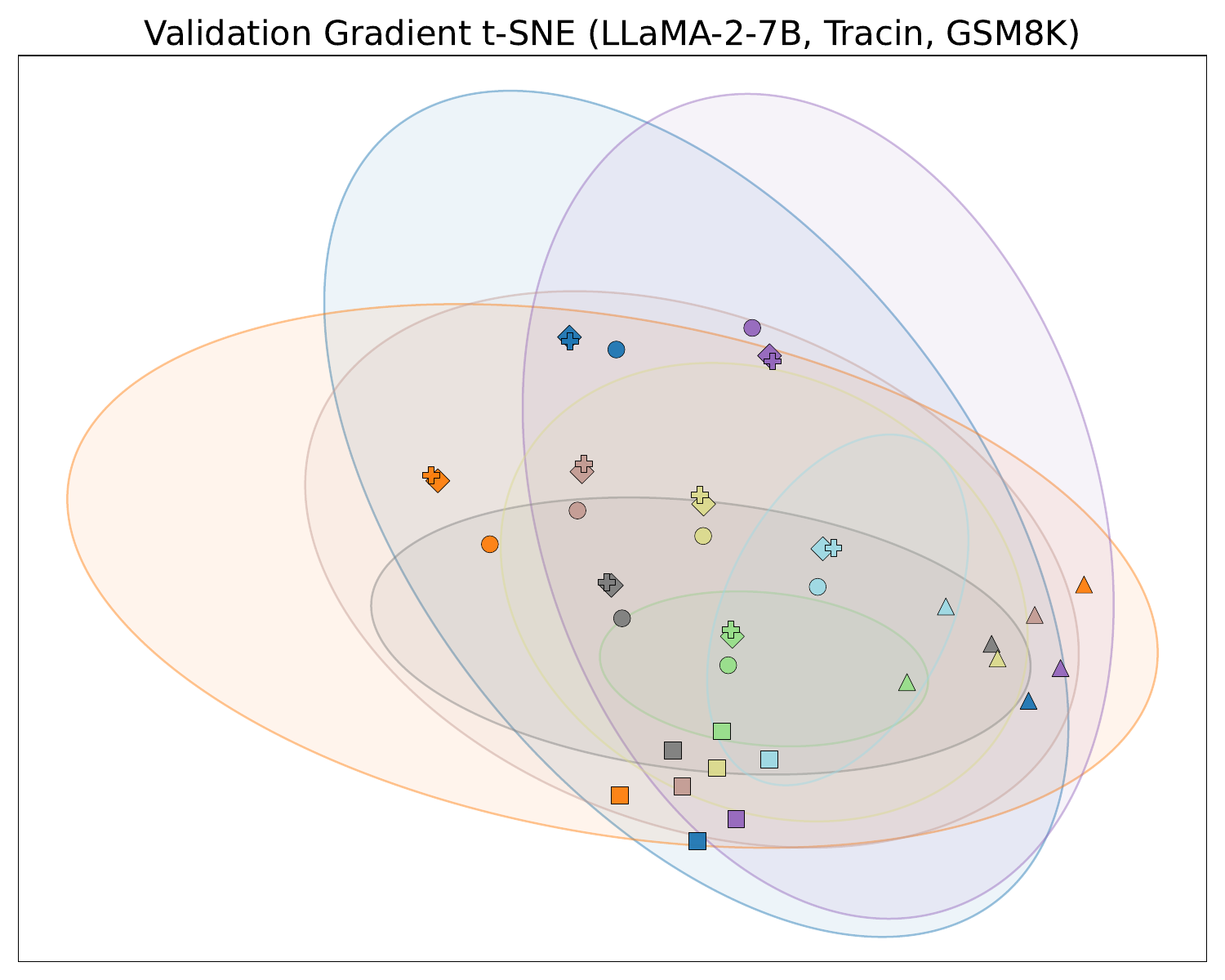}
  \includegraphics[width=0.33\linewidth]{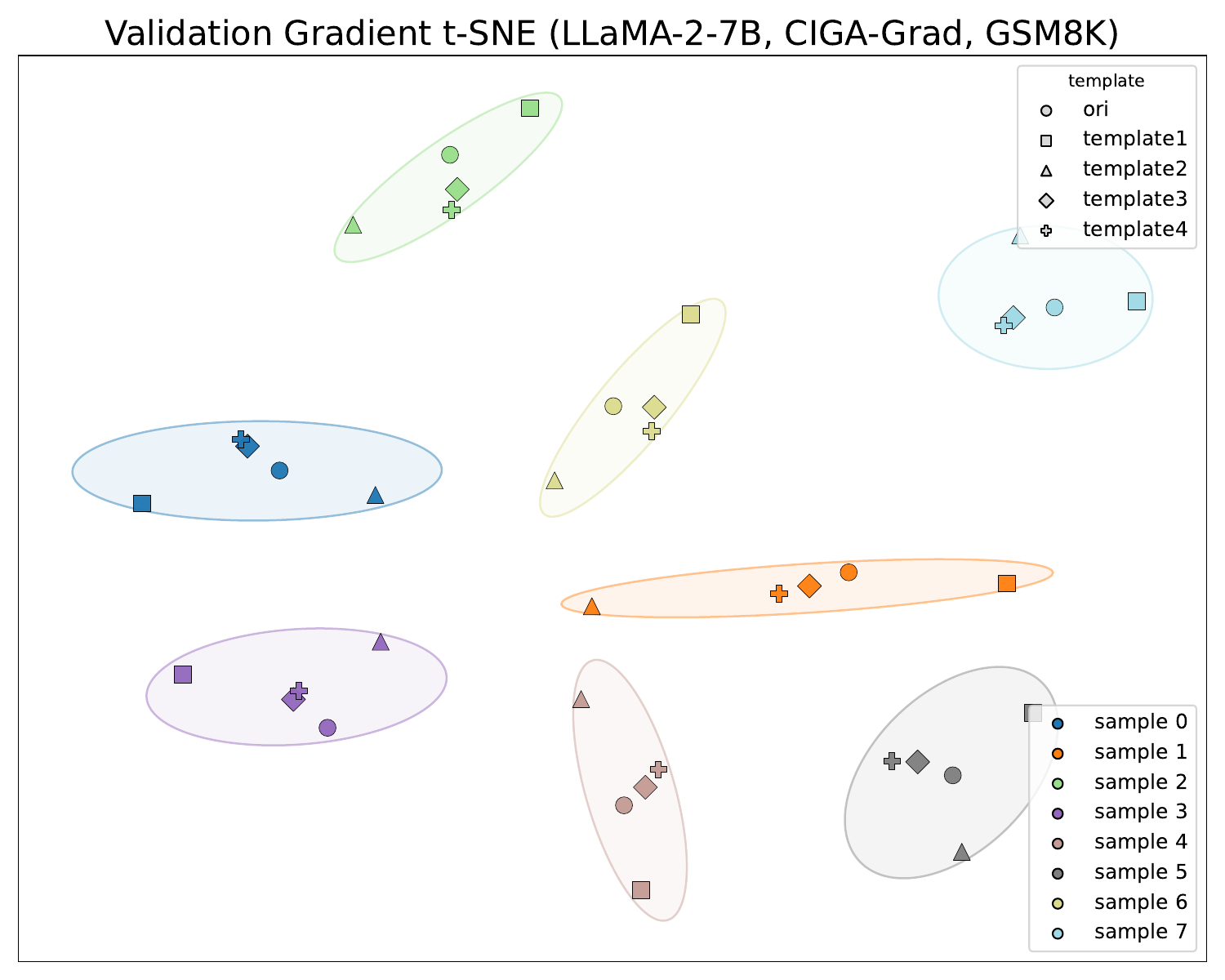}
  \caption{Comparison of gradient distributions via t-SNE.}
  \label{fig:tsne}
\end{figure*}

To investigate the surface form bias in gradient methods, we use t-SNE to visualize the gradient features of validation samples and their semantics-preserving surface-form variants. Figure \ref{fig:tsne} presents this distribution for LLaMA-2-7B on the GSM8K.

Observing global gradient methods reveals that different surface-form variants of the same validation data exhibit significant dispersion in the gradient space. This indicates that global gradient features are sensitive to semantically irrelevant variations in wording or formatting, leading to inconsistent parameter update directions for data sharing the same underlying content.

In contrast, the gradients extracted from the consistent influential layer by the CIGA framework can partially mitigate the interference of such noise, even without incorporating semantic similarity. The variants of the same underlying sample are clustered more tightly within the same neighborhood. This shows that extracting gradients from the consistent influential layer reduces the interference of surface forms and enables gradient attribution to focus more on the underlying content.

\subsection{Consistent Influential Score Analysis}
\begin{figure}[t]
  \includegraphics[width=\columnwidth]{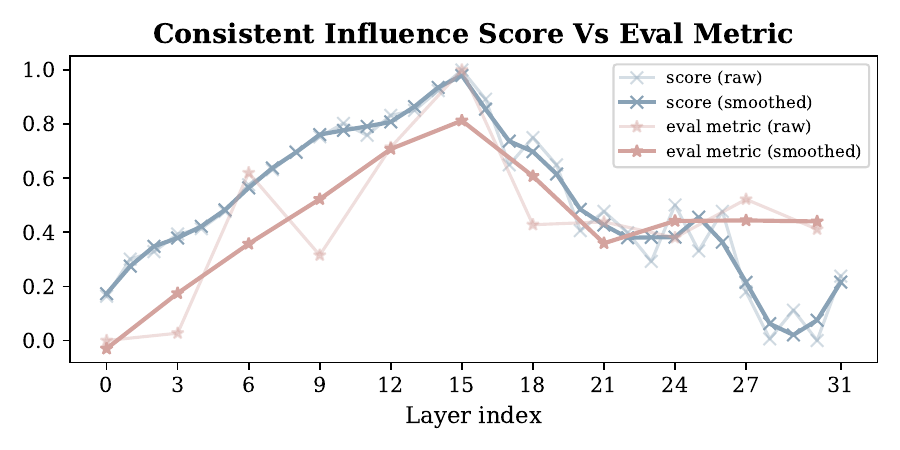}
  \caption{Relationship between the consistent influential score and downstream performance across layers.}
  \label{fig:layer-metrics}
\end{figure}

To examine whether the consistent influential score $S_l^{ci}$ can identify model layers suitable for gradient-based data attribution, we conduct a representative layer-wise analysis on MMLU with LLaMA-2-7B. Specifically, we sample one candidate every three layers, use the Grad-Score from each candidate layer to select a training subset under the same data budget, and then fine-tune and evaluate the model. Figure \ref{fig:layer-metrics} compares the normalized consistent influential score with downstream performance, showing both the raw curves and their smoothed trends.

\textbf{Alignment with Downstream Performance.} The global peak of the consistent influential score occurs around Layer 15 and falls within the middle-layer region with strong downstream performance. The two curves exhibit similar overall trends, indicating that $S_l^{ci}$, which jointly considers gradient strength and stability across surface forms, can identify layers whose gradient features are more suitable for data selection. This provides direct empirical support for using the score to determine the consistent influential layer.
    
\textbf{Preference for Middle Layers.} High consistent influential scores and strong downstream performance are both concentrated in the middle layers. This observation is consistent with the cross-task layer-selection results above: CIGA identifies a stable middle-layer region rather than relying on a fixed absolute layer index. One plausible explanation is that shallow layers encode more local lexical and syntactic features, while deeper layers are more closely tied to output format and generation distributions; middle layers can therefore provide a better balance among task-relevant information, gradient strength, and surface-form stability. Their gradient features consequently provide more effective attribution signals for data selection.

\section{Conclusion}

In this paper, we delve into the limitations of traditional global gradient attribution methods in data selection tasks for LLMs, and reveal the interference of surface form bias in data value evaluation. 
To overcome this challenge, we propose the CIGA framework. By combining a layer selection algorithm with dynamic feature fusion, CIGA integrates gradient-based attribution signals with the underlying semantic alignment features, achieving a complement between gradient utility and semantic features. 
Extensive experiments demonstrate that CIGA improves selection consistency under surface-form variations, while preserving downstream capabilities and achieving the highest average among data selection methods. CIGA attributes from one warm-up checkpoint and one selected layer, offering a more efficient alternative to multi-checkpoint global-gradient selection.

% In this paper, we identify surface-form bias in global-gradient data attribution: semantics-preserving changes to validation examples can alter gradient representations and lead to inconsistent selected subsets. To address this problem, we propose CIGA, which identifies a consistent influential layer by balancing update strength against cross-variant dispersion, and then combines selected-layer gradient alignment with task-adaptive semantic calibration for data ranking. Across two backbones and five benchmarks, CIGA substantially improves average selection consistency over global-gradient baselines while maintaining downstream performance. Ablation results show that consistent influential layer selection provides the main source of robustness, while semantic fusion offers complementary calibration of the gradient ranking. By performing attribution from one warm-up checkpoint and one selected layer, CIGA also reduces the cost of data selection relative to multi-checkpoint global-gradient methods. Overall, our findings show that separating content-relevant optimization signals from incidental surface-form variation provides a practical direction for more robust and efficient instruction data selection.

\section*{Limitations}

Although the CIGA framework demonstrates significant effectiveness and robustness in data selection for instruction fine-tuning, this study and the proposed method still exhibit the following limitations, which warrant further exploration and optimization in future work:

Inherent Computational Overhead: Despite CIGA substantially enhancing the noise robustness of gradients by precisely extracting consistent influential layers, as a gradient-based attribution framework, it inherently requires performing forward and backward passes across the entire candidate training corpus to compute and store the corresponding parameter gradients. When scaling to massive candidate data pools containing tens of millions of samples, this process incurs substantial temporal and computational overhead. Exploring methods to further compress gradient features or introduce efficient approximate computation techniques is crucial to alleviating this bottleneck.
    
Reliance on High-Quality Validation Sets: A core prerequisite for CIGA to accurately identify high-quality data is the utilization of a validation set from the target task to match core gradient and semantic features. While this mechanism endows the framework with robust task-specific adaptation capabilities, it also implies that for entirely novel target tasks lacking off-the-shelf validation sets, researchers or developers must incur additional costs to construct prior reference datasets. Investigating how to leverage unsupervised methods or rely on the model's intrinsic knowledge to evaluate the value of training data will be a primary focus of our future research.

% \section*{Acknowledgments}

% This document has been adapted
% by Steven Bethard, Ryan Cotterell and Rui Yan
% from the instructions for earlier ACL and NAACL proceedings, including those for
% ACL 2019 by Douwe Kiela and Ivan Vuli\'{c},
% NAACL 2019 by Stephanie Lukin and Alla Roskovskaya,
% ACL 2018 by Shay Cohen, Kevin Gimpel, and Wei Lu,
% NAACL 2018 by Margaret Mitchell and Stephanie Lukin,
% Bib\TeX{} suggestions for (NA)ACL 2017/2018 from Jason Eisner,
% ACL 2017 by Dan Gildea and Min-Yen Kan,
% NAACL 2017 by Margaret Mitchell,
% ACL 2012 by Maggie Li and Michael White,
% ACL 2010 by Jing-Shin Chang and Philipp Koehn,
% ACL 2008 by Johanna D. Moore, Simone Teufel, James Allan, and Sadaoki Furui,
% ACL 2005 by Hwee Tou Ng and Kemal Oflazer,
% ACL 2002 by Eugene Charniak and Dekang Lin,
% and earlier ACL and EACL formats written by several people, including
% John Chen, Henry S. Thompson and Donald Walker.
% Additional elements were taken from the formatting instructions of the \emph{International Joint Conference on Artificial Intelligence} and the \emph{Conference on Computer Vision and Pattern Recognition}.

\bibliography{custom}

\appendix

% \section{Appendix}
% \label{sec:appendix}
\section{Experimental Setup Details}
\subsection{Dataset Composition and Scale}
\label{sec:appendix_datasets}

To ensure our data selection framework is evaluated on a sufficiently diverse and massive candidate pool, our training corpus aggregates several widely adopted collections. Table \ref{tab:dataset_composition} summarizes the approximate volume and primary domain of each dataset subset used in our experiments.

\begin{table*}[h!]
  \centering
  \caption{Overview of the candidate training datasets, including their approximate volume and primary content focus.}
  \label{tab:dataset_composition}
  % \resizebox{\linewidth}{!}{
  \begin{tabular}{l | c l}
    \toprule
    \textbf{Dataset} & \textbf{Volume} & \textbf{Primary Content Domain} \\
    \midrule
    \textbf{FLAN v2} \citep{flanv2}   & 100000 & Zero shot and few shot NLP tasks \\
    \textbf{CoT} \citep{cot}          & 100000 & Step by step logical deduction \\
    \textbf{Alpaca GPT4} \citep{alpaca} & 52002 & High quality general instructions \\
    \textbf{Dolly} \citep{dolly}      & 15011 & Human generated conversational records \\
    \textbf{OASST1} \citep{oasst}     & 55668 & Multi turn dialogue sequences \\
    \textbf{GSM8K} \citep{gsm8k}      & 7473      & Grade school mathematical reasoning \\
    \bottomrule
  \end{tabular}
  % }
\end{table*}

These diverse data sources collectively contribute to the foundation models' general instruction following capability, complex reasoning skills, and conversational alignment.

Regarding the licenses and intended use of these artifacts, we confirm that all public datasets and base models were utilized strictly for their intended academic research purposes (i.e., instruction tuning and model evaluation) in compliance with their original licenses. Upon acceptance, we will publicly release the CIGA codebase. This release is strictly intended to facilitate non-commercial academic research, ensuring full compatibility with the original access conditions and intended use of the source artifacts.

\subsection{Detailed Baseline Descriptions}
\label{sec:appendix_baselines}

To comprehensively evaluate the effectiveness of the CIGA framework, we compare it against a diverse set of strong data selection baselines.

\textbf{Semantic Matching Methods.} 
This category relies on dense representation retrieval to find training samples semantically aligned with the validation set. We utilize the widely adopted all MiniLM-L6-v2 \citep{reimers-2019-sentence-bert} sentence transformer to encode both the training candidate pool and the validation sets into a shared embedding space. Data selection is subsequently performed by ranking the cosine similarity scores.

\textbf{Gradient Methods.} 
These methods attribute model performance directly to individual training samples by leveraging parameter gradients.
\begin{itemize}
\item \textbf{LESS} \citep{Xia2024LESSSI}: A state of the art gradient attribution method that projects global parameter gradients into a lower dimensional space using random projections. It computes the influence score by aggregating gradient similarities across multiple saved training checkpoints. We followed LESS's methods for gradient computation and saving, but utilized only a single checkpoint; consequently, our efficiency is significantly higher than that of LESS.

\item \textbf{TracIn} \citep{tracin}: A first-order gradient attribution method that traces the training trajectory. It scores a training example by the learning-rate-weighted sum, over multiple checkpoints, of the dot products between its gradient and the validation gradients, which estimates how much training on that example reduces the validation loss.
\end{itemize}

\textbf{Information based Methods.} 
Instead of relying on heavy gradient computations, these methods select data based on intrinsic model feedback, such as perplexity, confidence, or generation difficulty during forward pass inference.
\begin{itemize}
    \item \textbf{IFD} \citep{ifd}: Instruction Following Difficulty identifies training samples that exhibit a high ratio of condition answer loss to direct language modeling loss. It operates on the premise that samples inherently difficult for the model to generate directly, yet coherent given the instruction, yield the highest training utility.
    \item \textbf{SelectIT} \citep{liu2024selectit}: An advanced selection algorithm evaluating data based on instruction following difficulty and model confidence. It dynamically filters instruction tuning data by analyzing the foundation model's generation probabilities. Given the massive volume of our combined training corpus, we specifically reproduce the SelectIT model level score for the LLaMA-2-7B and LLaMA-2-13B architectures, discarding the LLaMA-2-70B score from the original paper.
\end{itemize}

% \subsection{Hardware Environment}
% \label{app:hard}
% All core experiments, including model fine-tuning and gradient computation, were conducted on an Ubuntu Linux server equipped with 4 NVIDIA A100 GPUs (80GB).

% \subsection{SNR}
% From a statistical perspective, our formal definition of gradient SNR is mathematically isomorphic to the F-statistic in Analysis of Variance (ANOVA). By treating the semantic variance across different data samples as inter-group variance (signal) and the variance induced by prompt words as intra-group error (noise), our SNR explicitly measures the statistical separability of features in the gradient space. This provides a rigorous theoretical foundation for its use as a fusion confidence metric: an extremely high SNR implies that gradient attribution possesses strong statistical robustness against prompt template perturbations.

\subsection{Evaluation Protocol and Prompt Variants}
\label{app:templates}

Selection consistency is the Jaccard similarity \citep{jaccard1912distribution} between Top-$k\%$ subsets obtained from different variants. With five variants, we average all $\binom{5}{2}=10$ pairwise values. Downstream performance is measured by TyDiQA F1, MMLU and HellaSwag accuracy, and BBH and GSM8K exact match after retraining on the selected subset.

For each target task, we manually construct five variants of the same validation examples. They preserve the task semantics and question/answer content while changing only semantically irrelevant wording or layout. CIGA, LESS, and TracIn use the same variants in the consistency evaluation. Table \ref{tab:prompt_templates} lists the exact templates.

\begin{table*}[htbp]
  \centering
  \small % 稍微缩小字号以适应长文本
  \renewcommand{\arraystretch}{1.3} % 增加行高，让多行文本看起来不拥挤
  \caption{Details of prompt variant templates across different benchmark datasets.}
  \label{tab:prompt_templates}
  \begin{tabular}{llp{0.7\linewidth}}
    \toprule
    \textbf{Dataset} & \textbf{Template} & \textbf{Content} \\
    \midrule
    \multirow{5}{*}{\textbf{TyDiQA}} 
    & Template 0 & Answer the following question based on the information in the given passage. Passage: \texttt{\{C\}}\textbackslash nQuestion: \texttt{\{Q\}}\textbackslash nAnswer: \texttt{\{A\}} \\
    & Template 1 & Read the text below and provide a answer to the query. Text: \texttt{\{C\}}\textbackslash nQ: \texttt{\{Q\}}\textbackslash nA: \texttt{\{A\}} \\
    & Template 2 & Given the background context, please extract the answer. Context: \texttt{\{C\}}\textbackslash nQuery: \texttt{\{Q\}}\textbackslash nResponse: \texttt{\{A\}} \\
    & Template 3 & Reference: \texttt{\{C\}}\textbackslash nQuestion: \texttt{\{Q\}}\textbackslash nResult: \texttt{\{A\}} \\
    & Template 4 & Answer the following questions. \texttt{\{C\}}\textbackslash n\texttt{\{Q\}}\textbackslash n\texttt{\{A\}} \\
    \midrule
    \multirow{6}{*}{\textbf{MMLU}} 
    & \textit{Full Format} & \texttt{\{prefix\}}\textbackslash n\textbackslash n\texttt{\{Q\}}\textbackslash nA. ...\textbackslash nB. ...\textbackslash nC. ...\textbackslash nD. ...\textbackslash nThe answer is: \texttt{\{A\}} \\
    & Template 0 & The following are multiple choice questions (with answers) about \texttt{\{S\}}. \\
    & Template 1 & Choose the correct answer (A, B, C, or D) for the question about \texttt{\{S\}}. \\
    & Template 2 & Multiple choice question on \texttt{\{S\}}: \\
    & Template 3 & Select the best option for the following question. \\
    & Template 4 & Answer the following question about \texttt{\{S\}}. \\
    \midrule
    \multirow{6}{*}{\textbf{BBH}} 
    & \textit{Full Format} & \texttt{\{task\_instruction\}}\textbackslash n\textbackslash n\texttt{\{Q\}}\textbackslash nA: \texttt{\{A\}} \\
    & Template 0 & \texttt{\{T\}} \\
    & Template 1 & Read the instructions, then answer the question below.\textbackslash n\textbackslash n\texttt{\{T\}} \\
    & Template 2 & Below is a reasoning task. Answer the final question.\textbackslash n\textbackslash n\texttt{\{T\}} \\
    & Template 3 & Answer the following reasoning question.\textbackslash n\texttt{\{T\}} \\
    & Template 4 & Answer the question below. \\
    \midrule
    \multirow{5}{*}{\textbf{GSM8K}} 
    & Template 0 & Solve the following math problem.\textbackslash n\textbackslash n\texttt{\{Q\}}\textbackslash nAnswer: \texttt{\{A\}} \\
    & Template 1 & Solve the math problem below.\textbackslash n\textbackslash n\texttt{\{Q\}}\textbackslash nSolution: \texttt{\{A\}} \\
    & Template 2 & Here is a math problem, give the answer:\textbackslash n\textbackslash n\texttt{\{Q\}}\textbackslash n \texttt{\{A\}} \\
    & Template 3 & Work through the problem and give the answer.\textbackslash n\textbackslash n\texttt{\{Q\}}\textbackslash nAnswer: \texttt{\{A\}} \\
    & Template 4 & Read the problem and give the answer.\textbackslash n\textbackslash n\texttt{\{Q\}}\textbackslash nAnswer: \texttt{\{A\}} \\
    \midrule
    \multirow{5}{*}{\textbf{HellaSwag}} 
    & Template 0 & \texttt{\{C\}} \texttt{\{A\}} \\
    & Template 1 & Read the context and complete it with the best ending.\textbackslash n\textbackslash n\texttt{\{C\}} \texttt{\{A\}} \\
    & Template 2 & Write the ending that best continues the passage.\textbackslash n\textbackslash n\texttt{\{C\}} \texttt{\{A\}} \\
    & Template 3 & Write the most natural next sentence.\textbackslash n\textbackslash n\texttt{\{C\}} \texttt{\{A\}} \\
    & Template 4 & Write the most likely continuation for the description.\textbackslash n\textbackslash n\texttt{\{C\}} \texttt{\{A\}} \\
    \bottomrule
  \end{tabular}
\end{table*}

In Table \ref{tab:prompt_templates}, \texttt{\{C\}} represents the context or passage; \texttt{\{Q\}} the question or math problem; \texttt{\{A\}} the answer or ending continuation; \texttt{\{S\}} the MMLU subject; and \texttt{\{T\}} the BBH task instruction. For multilingual TyDiQA, the table lists only the English formats; corresponding translations are applied according to the data language.

\subsection{Training Parameters}
\label{app:training_parameters}
We fine-tune all models with LoRA. The LoRA rank is set to 128, the scaling factor alpha is set to 512, and the LoRA dropout rate is 0.1. Training is performed for one epoch with a maximum sequence length of 2048. We use AdamW with a learning rate of 2e-5, linear learning-rate scheduling, a warmup ratio of 0.03, and no weight decay. The warm-up and selected subsets each contain 5\% of the candidate pool. We set the RRF smoothing constant $r=60$, the semantic-head size $K=1000$, and $\epsilon=10^{-8}$. After subset selection, each method retrains the base model once per run. All core experiments were conducted on an Ubuntu Linux server with 4 NVIDIA A100 GPUs (80GB).

\section{Hyperparameter Sensitivity Analysis}
\label{app:hyperparameter}
We analyze the sensitivity of CIGA to the variance penalty $\alpha$ and the SNR sensitivity $\gamma$. Figures \ref{fig:alpha} and \ref{fig:gamma} report the TyDiQA results, while Table \ref{tab:hyperparameters} provides the complete results across all five tasks.

\subsection{Layer Selection Hyperparameter}

In the CIGA layer selection mechanism, hyperparameter $\alpha$ penalizes layer gradient variance across surface-form variants.

\begin{figure}[htbp]
  \centering
  \includegraphics[width=0.9\linewidth]{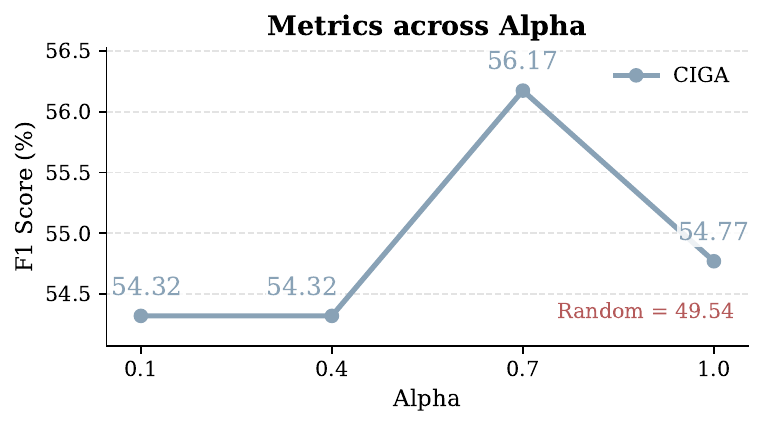}
  \caption{Sensitivity analysis of the variance penalty coefficient $\alpha$ in the CIGA layer selection mechanism.}
  \label{fig:alpha}
\end{figure}

As shown in Figure \ref{fig:alpha}, for small $\alpha$ (e.g., 0.1 and 0.4), performance stagnates around 54.3\%. This indicates that an insufficient variance penalty causes layer scores to rely overly on the gradient mean, leading the algorithm to select layers that are influential but more sensitive to format variations.
At $\alpha = 0.7$, the F1 score peaks near 56.2\%. This moderate penalty better balances update strength and format robustness, allowing CIGA to locate the consistent influential layer while filtering out format noise.
Conversely, increasing $\alpha$ to 1.0 degrades performance, as this excessive penalty can also suppress useful task-related gradient signals.

We therefore select $\alpha=0.7$ on TyDiQA and keep it fixed for the other tasks and both backbones. Table \ref{tab:hyperparameters} confirms this choice: on the other tasks, $\alpha=0.7$ also achieves the best or second-best performance in most cases, together with the highest macro-average.

\subsection{Feature Fusion Hyperparameter}

The hyperparameter $\gamma$ controls how strongly the task-level fusion weight $\beta$ responds to the gradient SNR.

\begin{figure}[htbp]
  \centering
  \includegraphics[width=0.9\linewidth]{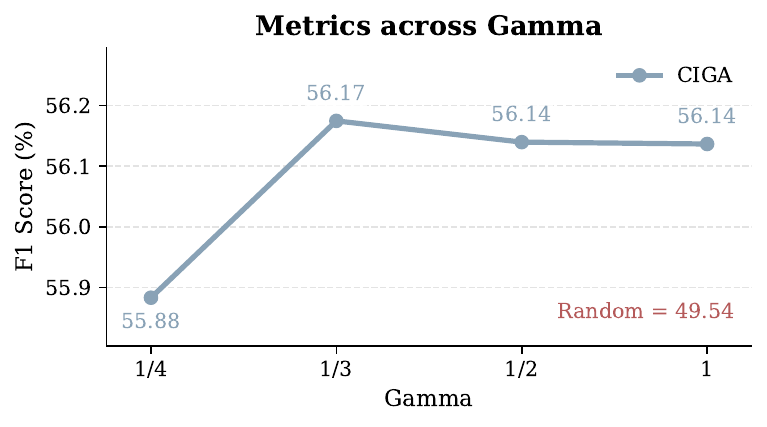}
  \caption{Sensitivity analysis of the SNR sensitivity $\gamma$ in dynamic feature fusion.}
  \label{fig:gamma}
\end{figure}

As shown in Figure \ref{fig:gamma}, when the value of $\gamma$ is greater than 1/4, performance remains within a narrow high-performing range. This indicates that a moderate SNR sensitivity keeps the algorithm more reliant on high-confidence gradient rankings rather than placing excessive weight on semantic similarity.
However, reducing $\gamma$ to $1/4$ degrades performance, as this excessively small value weakens the impact of high-confidence gradients and allows the semantic ranking to receive a larger fusion weight.

On TyDiQA, $\gamma=1/3$ gives the best F1, so we select it and keep it fixed for the other tasks and both backbones. Table \ref{tab:hyperparameters} confirms that this choice also yields the highest macro-average across the five tasks.

\begin{table*}[htbp] % 如果单栏能够放下，可以去掉星号变成 \begin{table}[htbp]
  \centering
  \caption{Performance metrics (\%) under different configurations of hyperparameter $\alpha$ and $\gamma$.}
  \label{tab:hyperparameters}
  \begin{tabular}{llcccccc}
    \toprule
    \textbf{Hyperparameter} & \textbf{Value} & \textbf{TydiQA} & \textbf{MMLU} & \textbf{BBH} & \textbf{GSM8K} & \textbf{HellaSwag} & \textbf{Avg} \\
    \midrule
    \multirow{4}{*}{Variance Penalty ($\alpha$)} 
    & 0.1 & 54.32 & 45.77 & 41.20 & 17.66 & \textbf{74.62} & 46.71 \\
    & 0.4 & 54.32 & \textbf{47.46} & 41.20 & 17.66 & 74.54 & 47.04 \\
    & 0.7 & \textbf{56.17} & \textbf{47.46} & \textbf{41.39} & \textbf{19.64} & 74.54 & \textbf{47.84} \\
    & 1.0 & 54.77 & \textbf{47.46} & 37.78 & \textbf{19.64} & 74.54 & 46.84 \\
    \midrule
    \multirow{4}{*}{Fusion Coefficient ($\gamma$)} 
    & 1   & 56.14 & 47.45 & 41.20 & 19.18 & 74.47 & 47.69 \\
    & 1/2 & 56.14 & 47.29 & 40.93 & 19.11 & 74.41 & 47.58 \\
    & 1/3 & \textbf{56.17} & \textbf{47.46} & \textbf{41.39} & \textbf{19.64} & \textbf{74.54} & \textbf{47.84} \\
    & 1/4 & 55.88 & 47.37 & 40.56 & 19.41 & 74.46 & 47.54 \\
    \bottomrule
  \end{tabular}
\end{table*}

\subsection{Selection Ratio}
\label{app:selection_ratio}
\begin{figure*}[htbp]
  \centering
  \includegraphics[width=\linewidth]{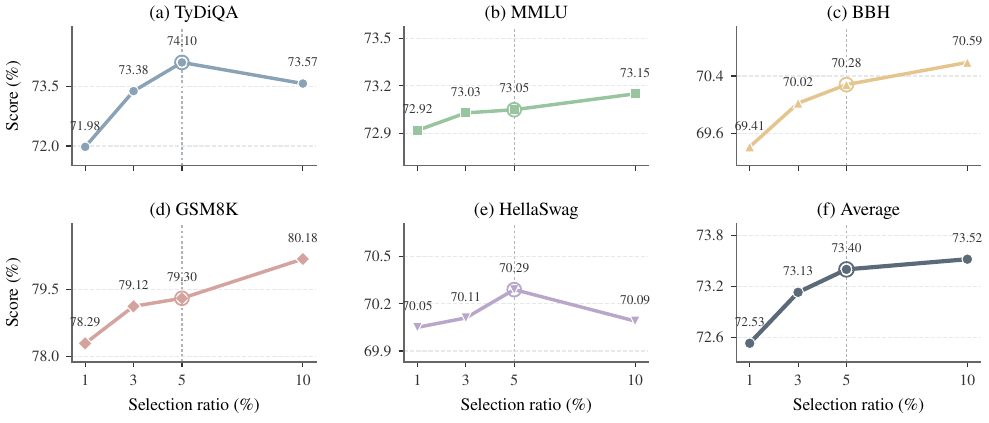}
  \caption{Selection-ratio ablation on Qwen3-4B-Base. Scores are reported in percent. The dotted line marks the default 5\% ratio; panels use task-specific y-axis ranges.}
  \label{fig:ratio}
\end{figure*}

We vary the selection ratio from 1\% to 10\% on Qwen3-4B-Base to evaluate CIGA's sensitivity to the data budget. As shown in Figure \ref{fig:ratio}, the macro-average remains within 72.53\%--73.52\% and generally improves as the data budget increases.

\section{Declaration of Generative AI Tool Usage}
During the preparation of this manuscript, the authors utilized LLMs
to assist with several auxiliary tasks. The primary applications of the AI tools included: Refining the grammar, phrasing, and flow of the English text, as well as translating non-English data samples, to improve overall readability. Structuring and formatting the LaTeX table for the clear presentation of the case study.

The generative AI tools were not used to conceptualize or generate the core scientific contributions, methodology, or theoretical claims of this work. All experimental designs, data analyses, and final scientific conclusions represent the original work of the authors. The authors have rigorously reviewed, edited, and verified all AI-assisted content and take full responsibility for the accuracy and integrity of this publication.

\section{Case Study}
We present a detailed case study on the BBH dataset (Table \ref{tab:case_study_bbh}). The evaluation instance tests logical reasoning via a step-by-step evaluation of a Boolean expression. As shown, CIGA retrieves a single-turn step-by-step reasoning sample, directly aligning with the logical deduction requirements of the target task. In contrast, baseline methods struggle with structural alignment: Cos selects an irrelevant instruction-following task about "steps", LESS retrieves a noisy multi-turn dialogue containing error corrections, and TracIn selects an answer that completely lacks the necessary reasoning steps.

\begin{table*}[t]
  \centering
  \caption{Top-1 selected training samples by different methods for the BBH evaluation task.}
  \label{tab:case_study_bbh}
  \begin{tabular}{l | p{13.5cm}}
    \toprule
    \textbf{Method} & \textbf{Text (Validation Example / Selected Training Sample)} \\
    \midrule
    \textbf{Eval Data} & 
    \textit{BBH (Boolean Expressions)}:\newline 
    [User]: Evaluate the result of a random Boolean expression.\newline 
    Q: not ( ( not not True ) ) is \newline 
    [Assistant]: Let's think step by step. Remember that (i) expressions inside brackets are always evaluated first and that (ii) the order of operations from highest priority to lowest priority is "not", "and", "or", respectively. We first simplify this expression "Z" as follows... \textcolor{gray}{\textit{[... Deduction steps truncated ...]}} ...So the answer is False. \\
    \midrule
    \midrule
    \textbf{Cos} & 
    \textit{dolly\_2928}:\newline 
    [User]: Given these two sets of instructions, will you end up at the same end point if you started at the same beginning point? First instructions: 1) Take 2 steps forward... \newline 
    [Assistant]: Yes, you will end up at the same point for both sets of instructions... \\
    \midrule
    \textbf{LESS} & 
    \textit{oasst1\_15234}:\newline 
    [User]: If you have a 3 gallon cylindrical shaped jug and a 5 gallon... \newline 
    [Assistant]: The correct procedure to end up with exactly 4 gallons in the 5 gallon jug is: b... \newline 
    [User]: i dont understand, can you explain it to me like I was like 5? \newline 
    [Assistant]: My apologies if that was confusing. I have realized that my previous answer was incorrect... \\
    \midrule
    \textbf{TracIn} & 
    \textit{oasst1\_15229}:\newline 
    [User]: If you have a 3 gallon cylindrical shaped jug and a 5 gallon... \newline 
    [Assistant]: The correct answer is: \textbf{``e. None of the above''}. \\
    \midrule
    \textbf{CIGA} & 
    \textit{oasst1\_15233}:\newline 
    [User]: If you have a 3 gallon cylindrical shaped jug and a 5 gallon... \newline 
    [Assistant]: The correct procedure to end up with exactly 4 gallons in the 5 gallon jug is: \newline 
    b. Fill 3 gallon jug. Pour it all into the 5 gallon jug twice \newline 
    Here's the explanation: \newline 
    1. Fill the 3 gallon jug with water. \newline 
    2. Pour all 3 gallons from the 3 gallon jug into the 5 gallon jug... \\
    \bottomrule
  \end{tabular}
\end{table*}

\end{document}